\pdfoutput=1
\documentclass{article}

\usepackage{arxiv}

\usepackage[utf8]{inputenc} 
\usepackage[T1]{fontenc}    
\usepackage{hyperref}       
\usepackage{url}            
\usepackage{booktabs}       
\usepackage{amsfonts}       
\usepackage{nicefrac}       
\usepackage{microtype}      
\usepackage{lipsum}

\usepackage{pgfplots}
\pgfplotsset{width=8cm, compat=1.14}
\usepackage{amsmath}
\usepackage{algorithm}
\usepackage{algpseudocode}
\usepackage{tikz}
\usepackage{iftex}
\ifPDFTeX
\usepackage{textcomp}
\fi
\usetikzlibrary{angles,quotes}
\DeclareMathOperator*{\argmin}{argmin}

\title{
Hybrid Machine Learning Approach to Popularity Prediction of Newly Released Contents for Online Video Streaming Service
}

\author{
  Hongjun Jeon \\
  TEAMLAB \\
  Gachon University\\
  Seongnam, Repulbic of Korea \\
  \texttt{jeonhongjun730@gmail.com} \\
 \And
 Wonchul Seo\\
 Graduate School of Management of Technology\\
 Pukyong National University\\
 Busan, Repulic of Korea \\
 \texttt{wcseo@pknu.ac.kr} \\
\And
Eunjeong Lucy Park\thanks{They are both co-corresponing authors. Tel: +82-31-750-5804, Email: teamlab.gachon@gmail.com, Web: theteamlab.io} \\
Papago \\
NAVER \\
Seongnam,  Repulbic of Korea  \\
\texttt{lucy.park@navercorp.com} \\
\And
 Sungchul Choi\footnote[1]{}  \\
  Department of Industrial and Management Engineering\\
  TEAMLAB at Gachon University\\
  Seongnam, Repulbic of Korea \\
  \texttt{sc82.choi@gachon.ac.kr} \\
}

\begin{document}
\maketitle

\begin{abstract}
In the industry of video content providers such as VOD and IPTV, predicting the popularity of video contents in advance is critical not only from a marketing perspective but also from a network optimization perspective.  By prognosticating successful preference of user content, its large file will be efficiently deployed in the proper service providing server and ultimately lead to network cost optimization. Many previous studies have deployed view count prediction research to do this. However, the studies have been making predictions based on historical view count data from users . In cases such as these, contents had been published to the users and already sent to the service server. These approaches make it possible to efficiently deploy  content already published but are impossible to use for content that is not be published.  To address the problems, this research proposes a hybrid machine learning approach to the classification model for the popularity prediction of newly video contents which is not published. 

In this paper, we create a new variable based on the related content of the specific content and divide the entire dataset by the characteristics of the contents. Next, the prediction is performed using XGBoosting and deep neural net based model according to the data characteristics of the cluster. Our model uses metadata for contents for prediction, thus using categorical embedding techniques to solve the sparsity of categorical variables and requiring the system to learn efficiently for the specified deep neural net model. In addition, we use the FTRL-proximal algorithm to solve the problem of the view-count volatility of video content. The results would display overall better performance than the previous standalone method with a dataset from one of the top streaming service company.
\end{abstract}

\keywords{Streaming Service\and Popularity Prediction \and Embeddings \and Deep Learning \and Boosting Decision Tree}

\section{Introduction}

Along with the rise of online video streaming service market, the way in which video contents are consumed is changing drastically. Consumers are moving rapidly from traditional terrestrial broadcasting consumption to IPTV and online streaming service services, for example, YouTube and Netflix\footnote{Here's a guide to every live streaming service on the market or coming soon - https://read.bi/2VgcTb1}.  Recently, a new report from the Video Advertising Bureau\footnote{https://www.thevab.com/about-vab/} shows that the number of households that use only streaming services has tripled since 2013\footnote{The number of cord-cutters has tripled in the last 5 years, and it’s starting to hurt the TV channels - https://bit.ly/2EQacax}. According to the Global Internet Phenomena Report\footnote{https://www.sandvine.com/blog}, the massive online streaming platform currently accounts for 50.31\% of North American Internet traffic\footnote{Half of All Internet Traffic Goes to Netflix and YouTube - https://bit.ly/2Q6r52A}.

These changes in the video content market have increased service traffic at a rapid pace. Also, the rapid development of High-Definition(HD), Full-HD, 4K or more and 3D video, the large data size have caused additional service traffic issues. As a result, the streaming service company have faced an emerging need to actively respond to the change of the market regarding a system architecture for well-organized services. Due to the characteristics of streaming service, traffic monopoly by popular content is gaining in popularity, and this increase in traffic is a big issue for streaming service companies\footnote{https://pc.video.dmkt-sp.jp}. Accordingly, the development of a Content Delivery Networks (CDNs) that can actively address the problem has been increasingly in demand~\cite{7047922}.

To provide quality network and efficiency in streaming service, the service provider needs technology that pre-categorizes popular contents that are heavily consumed, all the while, storing them in storage on the fastest network available.  This process improves the capability of CDNs possible to react by providing huge consumer demand for high popular contents. To address problems of network efficiency, an online streaming platform arranges contents on storage by utilizing view-count records based on log data accumulated for a certain period after new content had been published. This method requires view-count records to consider time series characteristics, thus making it difficult to configure the content arrangement for efficient traffic management in one to two weeks. When new content is published, the content manager could arrange them through a heuristic approach but is problematic due to the immense amount of contents published every week.
 
The purpose of our research is to propose a classification model of the degree of popularity of new content in advance without historical view-count logs for the content.  The proposed model aims to provide efficient network management for service companies by providing a machine learning based framework classifying popular contents. We expect that companies can benefit from cost savings through network optimization by utilizing the proposed model.

In this paper, we present two contributions for the research of the content popularity prediction. First, hybrid approach is proposed to build machine learning model to predict the popularity of newly released contents.  Generally, the video contents could be separated into two types of contents having successive previous works or not having them. For example, a seasonal drama or weekly talk show will have previous related video contents. Whereas, movies or standalone documentaries sometimes would not have any other related previous works. The popularity of the previous content is an important feature to predict the popularity of new content. Second, we adopt batch FTRL-proximal optimizer algorithm~\cite{McMahan:2013:ACP:2487575.2488200} to address the volatility problem to train view-count log data. Basically, the demand of video streaming contents have higher volatility.  Subsequently, the demand of contents are high at the beginning of the release, but rapidly declines after a certain period of time. The view-count volatility affects the  gradient descent based training process, and distorts the performance of the model. Taking into consideration the characteristics of newly released contents, we propose the hybrid approach, based on XGBoost~\cite{chen2016xgboost} and Neural Net with batch FTRL-proximal optimizer algorithm~\cite{McMahan:2013:ACP:2487575.2488200}, demonstrating overall better performance than previous standalone method.

The rest of this paper is organized as follows. In Chapter 2, we review the related works of users’ content preference prediction and the popularity of media content prediction based on machine learning. In Chapter 3, we describe the proposed content prediction model and algorithms for the newly published contents without history log data. Based on this, In Chapter 4, we compare the performance of the proposed model and discuss our main contributions. Finally, we discuss conclusions and future works in Chapter 5.


\section{Related Works}
The researches for a popularity prediction method of contents have been widely conducted in various areas such as news, advertising, TV shows, streaming videos, and movie markets. The researches require the user's information, historical usage data, and metadata for contents. In reference to video contents, many recent researches utilize time-series based log data with content metadata, or external information such as text data written on social network applications.

Initially, many researchers focused on the analysis of tree series based on meta data of contents and QOE data. Nielsen, Sereday and Cui used XGBoosting~\cite{DBLP:journals/corr/ChenG16} to predict the audience ratings of TV programs in United States over the next year~\cite{MEDIA}. More recently, researchers have further considered clustering techniques for data before view count prediction. Chengang Zhu and Guang Cheng used Chinese online streaming data to predict the popularity of the program using the Random Forest model after clustering trend using Dynamic Time Wrapping (DTW) algorithm and K-medoids clustering~\cite{8086153}. For newly imported programs, thy used GBM Classification to allocate them to existing program trends. A similar study of program prediction using clustering is the case of estimating TV popularity rating using incremental k-means clustering proposed by D. Anand and A.V.Satyavani~\cite{kmeans}. Incremental k-means clustering is a technique to register new content when it comes into the database and grouping it with existing content clusters that already exist. This is different from the previous k-medoids clustering~\cite{8086153}, in which there is a difference in the clustering method for new programs entering in the middle of the process.
 
Another research used the text data for the contents to perform view count prediction. Fukushima and Yusuke proposed the model with a focus on variables for actors, staff, and directors rather than on the program's metadata or watching records to predict the popularity of contents~\cite{7545038}. They also used it as a predictor of the model, taking into account how much of the actor was mentioned in social media and how many awards there were. The prediction model was done using Support Vector Regression (SVR). In the field of movies similar to streaming contents, Mestyan et al. predicted the movie's success and financial success based on the number of editors contributing to the Wikipedia article on cinema and movie audiences~\cite{10.1371/journal.pone.0071226}. Similarly, to predict the count of the viewing data, it is vital to quantify various external variables in addition to the primary data about the contents to improve the performance.

However, these studies are somewhat different from our proposed model where view-count predictions are performed at the point of when the content is not published in that we have used a time series model based on past view count of specific content. As a result, the methods of previous studies may have a profound effect on the performance improvement of broadcast program popularity estimation using the time series of the content.  However, before these time series data are accumulated, it is hard to predict the popularity of newly released contents and storage optimizer to perform properly for them. Consequently, the scarcity of research focusing on predicting popularity of the newly released contents are not consistent with previous view-count history data.

\section{Framework}
\subsection{Problem Formulation}

In this research, we propose the model to predict popularity of new content that has not been released yet by using information of already released content. We can define the problem as follows. First, $c^{t}$ is an individual content, an instance of data at present time {t}. We define the content $c_{new}^{t} \notin C^{t}$ that has not yet been published and content $c_{old}^{t}$ $\in$ $C^{t}$ that has already been released from a set of all contents $C$ that are observed from a period $r$ to present time {t}. Next, we build a model predicting $P_{c_{new}^{t}}(c_{new}^{t}, C^{t})$ which is a probability representing whether $c_{new}^{t}$ is a high popular content or not. In $P_{c_{new}^{t}}(c_{new}^{t}, C^{t})$, $C^{t}$ is the view-count history log data set including the their metadata at the present time $t$, and $c_{new}^{t}$ is a set of metadata of new content. After predicting the possibility, the set of $c_{new}$ is converted to $c_{old}$ at the next $t+1$ sequence after being predicted. In the end, the better the prediction, the closer the predicted probability $P_{c_{new}}(t-r, t)$ with a scale of [0,1] and the actual value $P_{c_{old} \in c_{new}}(t-r +1, t+1)$ which is a binary value $1$ or $0$. $1$ is high popular content and $0$ is not popular content. In this paper, we designate them as hot content and cold content, respectively.

\subsection{Method Overview}

Our model is divided into three stages. First, we collect historical log data and content metadata from the streaming service database. In this stage, the set of contents are separated into two types: Contents having successive previous works and not having them. In this paper, we call them A type and B type respectively. The proposed hybrid model uses different features for each type. In A type, because it is available to utilize the data regarding view-count of previous contents, we comprise structured dataset including the data of previous contents. However, in B type, due to the fact that related data could not be available, we utilize additional metadata such as unstructured data. Metadata of the video contents includes many kinds of text data, for example actor names and keywords for content. By adding the additional information, we complement the shortage of information to B type contents. The overall utilized features are described in Table~\ref{tab:a}.

\begin{table}[!h]
\centering

\caption{Descriptions of Attributes}
\label{tab:a}
\renewcommand{\arraystretch}{1.3}
\begin{tabular}{l l l l}

\hline
ID&
Feature& 
Description& 
Example \\
\hline
1&
Payment& 
Payment method& 
Pay, Free\\
2&
Type& 
Program Type& 
Drama, Movie, Ani \\
3&
Genre& 
Genre Details& 
Mystery, Romance \\
4&
Playtime& 
Regeneration time& 
3072, 5047(s) \\
5&
Episode& 
episodes count& 
1, 25, 45 \\
6&
Age& 
an age limit& 
19,17,15 \\
7&
Channel& 
Program channel& 
TBS, NTV \\
8&
Actor& 
actor, Voice actor& 
Sung Dong Il,  Seo Yu ri \\
9&
title& 
episode title& 
Detective Conan 7th \\
10&
Keyword& 
episode Keyword& 
Japan, famous actor \\
11&
Release& 
publish date& 
20180716 \\
12&
Related view& 
related works rating& 
40728, 6587 \\
\hline
\end{tabular}
\end{table}

The second stage is training the dataset with two different models. Type A  dataset only has structured features, and Type B dataset has structured features and unstructured dataset. Because of the characteristics of each dataset, we apply the tree-based model and neural net based model to two different datasets, individually. XGBoosting model is used to type A dataset. XGBoosting is a tree-based model utilizing a technique called Boosting. Boosting is a technique for predicting accuracy by combining weak classifiers into sets, and XGBoosting is an intuitive model characterized by rapid learning and classification based on parallel processing. Recently, this model is widely used and proved to be the most accurate and useful technique for structured dataset through the Data Competition. Next, because a tree-based model is not suitable to unstructured data like text, we utilize the neural network with categorical embedding techniques for type B. Recently, the deep learning, which is another name of the neural network, is a best suitable algorithm for unstructured data such as text. We use embedding techniques representing the compressed representation of learning on latent space of data. We achieve overall better performance than a standalone model by composing the hybrid model.

\begin{figure}[!h]
\centering
\includegraphics[width=0.5\columnwidth]{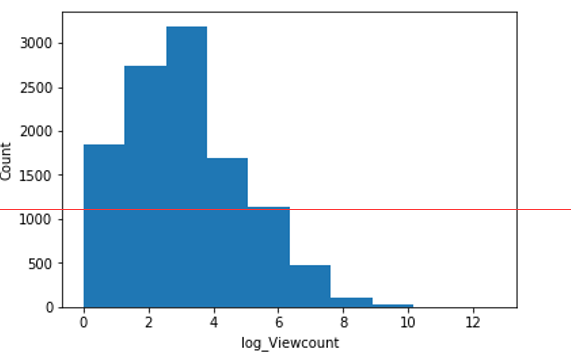}
\caption{Relationship between content count and view count.}
\label{fig:a}
\end{figure}

The final stage is to measure the performance of the proposed model. The central purpose of the proposed model is to provide information regarding whether the popularity of content is high or not for supporting to deploy large file of contents to CDNs for efficient network management. The demand for stream service typically shows long-tail distribution, which means some popular contents uses the most network resources. As shown in Fig~\ref{fig:a}, the only few contents, referred to as hot contents, have high view-count, and others, referred to as cold contents, have very low view-count. The distribution of recent contents determines the criterion for dividing the popular contents of a label. We set the boundary as top 20\% to separate hot and cold contents for a network optimization point of view for storage distribution. Based on the boundary, we train the model to predict the classification results of hot and cold contents which are not released. To compare the performance of the model, we use recall, precision, and f1-score as performance evaluation metrics. Because the dataset is imbalanced, we focus on the accurate prediction for the classification of hot contents.  Also, since the dataset not considering past view count does not exist, The popular content in some duration is due to the performance value being 0. Thus we calculate the overall average of classification of results during all periods.

The overall training algorithm of the framework is depicted in Algorithm 1.

\newcommand\tab[1][1cm]{\hspace*{#1}}
\newcommand{\factorial}{\ensuremath{\mbox{\sc Factorial}}}
\begin{algorithm}
\caption{Training Algorithm}\label{euclid}
\algrenewcommand\algorithmicrequire{\textbf{Input:}}
\algrenewcommand\algorithmicensure{\textbf{Output:}}

\begin{algorithmic}[1]
\Require 
	\Statex Structural dataset: $ \{ S_{1}, S_{2}, \cdots , S_{n} \} $
	\Statex Unstructural dataset: $ \{ T_{1}, T_{2}, \cdots , T_{n} \} $
    \Statex Length for used period of characteristic $A$: $r_{{}_{A}}$
    \Statex Length for used period of characteristic $B$: $r_{{}_{B}}$
\Ensure 
	\Statex {Ensemble model with learned parameters} \\
           {Initialize the training instance set $U$  $\leftarrow 0$ \; } 

\If {Program Type is characteristic $A$}
	\For{\textit{all available time interval} $t(1 \le t \le r_{{}_{A}})$}
		\State {Put ($S_{t}$) into $U_{A}$}
    \EndFor
\Else
	\For{\textit{all available time interval} $t(1 \le t \le r_{{}_{B}})$}
    	\State $H_{t}$ = Concat($S_{t}, T_{t}$)
        \State Put ($H_{t}$) into $U_{B}$
	\EndFor
\EndIf \\
{Initialize all the learnable parameters}\\
\textbf{Repeat}\\
\quad {Minimize the objective function within $U_{A}$ as Boosting}\\
\quad {Minimize the objective function within $U_{B}$ as NN}\\
\textbf{until} \emph{Convergence criterion met}\\
\textbf{Return} learned Ensemble model
\end{algorithmic}
\end{algorithm}

\subsection{Deep Neural Network Model using categorical embedding}

The type A dataset is typically structured data, named as tabular data, which if fitted to gradient based boosting algorithm.  However, type B dataset includes text dataset which is  unstructured and not fitted to the boosting algorithm. Therefore, we adopt deep neural net model to handle the type B dataset with an embedding technique.

The categorical data type is one obstacle to apply the neural net model. Because a categorical dataset generally is represented as a one-hot encoding vector. However the approach makes the dimensions of the input data larger, and the learning speed can also be lowered. As well, this cannot represent the relationship and semantic distance of each category.  To address the problem, the recent neural net researches utilize the embedding techniques representing the learning representation of a data by finding latent space. For example, a day of the week is an explanatory categorical variable in the view-count dataset. However, if we see the proper embedding representation, the variable can be defined as follows. 

\begin{equation}
Sunday : [0.8, 0.2, 0.1, 0.1], \\
Monday : [0.1, 0.2, 0.9, 0.9], \\
Tuesday : [0.2, 0.1, 0.9, 0.8]
\end{equation}

Monday and Tuesday are quite similar, but they are quite different from Sunday. The proposed neural network would learn the best representations for each category while it is training, and each dimension or direction, which doesn't necessarily line up with ordinal dimensions could have multiple meanings. In this research, we use Continuous Bag-of-Words (CBOW), which is introduced in the representative embedding technique called Word2Vec~\cite{DBLP:journals/corr/abs-1301-3781}, to build embedding space for categorical variables. 

To accomplish this, the following process is performed. $x_i$  is a one-hot encoding vector, the output of the layer of linear neurons given the input$x_i$  is defined as \begin{equation} X_i = \sum_{\alpha} \omega_{\alpha\beta}\delta_{x_{i}\beta} \end{equation}. $X_i$ is the value converted from discrete variable to vector.   If $m_i$ is the number of values for the categorical variable $x_i$ , $\delta_{x_i}\alpha$ then  is a vector of length $m_{i}$, where the element is only non-zero when $\alpha$ = $x_i$. $\omega_{\alpha\beta}$ is the weight connecting the one-hot encoding layer to the embedding layer and $\beta$ is the index of the embedding layer. The possible values for $\alpha$ are the same as $x_i$.

\begin{figure}
\centering
\includegraphics[width=0.8\columnwidth]{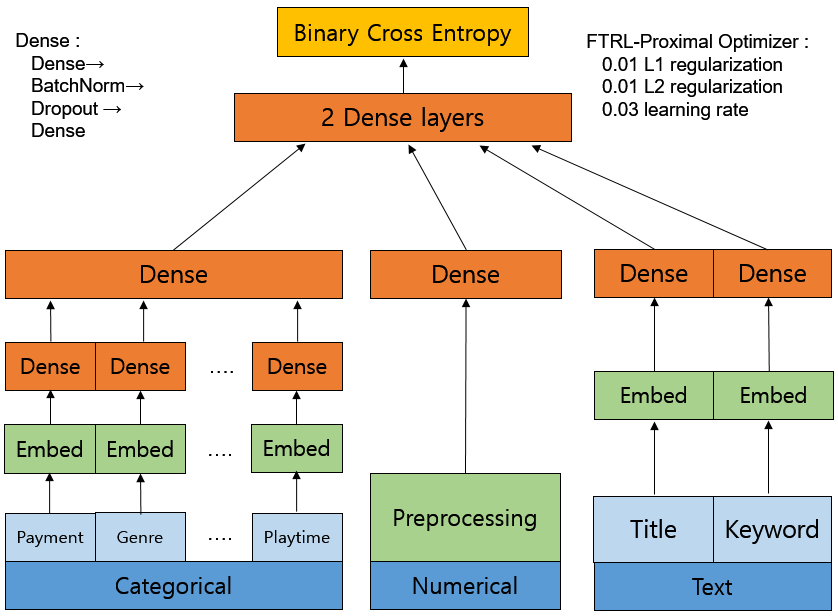}
\caption{Overview of the deep neural net with categorical embedding.}
\label{fig:b}
\end{figure}

We stack the deep neural net model using embedding vectors which is converted from categorical and continuous variables. The proposed model uses text data to learn both structured and unstructured data from the same layers. We learn all the categorical, numerical, and text data in one model. It is shown in Fig~\ref{fig:b}. First, each categorical variable is converted into a one-hot encoding and then entered independently of the input of the neural net. As you can see in Fig~\ref{fig:b}, the encoded input is entered independently of the the neural net and the values obtained after passing through the embedding and dense layer are concatenate. Based on seven categorical variables, three numeric variables, and two text data, we learn the model through the neural net considering all these features. We use a sigmoid function activation as an activation function of the output layer. Because there is no proper method for determining the number of dimensions of embedding when categorical variables are embedded, we have to find the suitable hyper parameter for them through empirical method.
 
\subsection{Optimizer}

Because the view-count of online-streaming contents have high volatility and sparsity, there are some difficulties to adopt gradient-based learning algorithm to massive scale dataset for online-streaming contents. In this research, the FTRL-Proximal algorithm~\cite{McMahan:2013:ACP:2487575.2488200} was employed to induce sparsity and volatility and it shows better performance. FTRL-Proximal algorithm is proposed by McMahan and H. Brendan. The algorithm is the analysis technique that provides the fastest, most real time analytical results, rather than performance, such as search engine-based ad prediction using the FTRL-Proximal algorithm published by Google. Like Google, search engine search history shows the popularity of certain words, which can be used effectively in view count prediction. The popularity of certain words in search engine also has high volatility.

Although the original algorithm is learned as online gradient descent, we change it to learned in batches for accuracy. The data we used are not as vast as Google search history data, and volatilities of data are relatively smaller than Google’s. Moreover, our study requires batch learning rather than online gradient descent because the purpose of our research requires higher accuracy than quick learning. Therefore, our study approaches the optimization method of this paper slightly different from the real-time learning method such as online gradient descent ~\cite{Bartlett:2007:AOG:2981562.2981571}. The modified FTRL updates the weights in the following manner. 

\begin{equation}
\mathbf{w}_{t+1}=\argmin_\mathbf{w} (\mathbf{g}_{1:t}\cdot \mathbf{w} + \frac{1}{2}\sum_{s=1}^t \sigma_{s}||\mathbf{w}-\mathbf{w}_{s}||^2_2+\lambda_1 ||\mathbf{w}||_1), 
\end{equation}

This algorithm defines $\sigma_{s}$ in terms of the learning rate schedule such that $\sigma_{1:t} = \frac{1}{\eta_{t}}$. $t$ is the current training batch and $\mathbf{g}_{t}$ is  a vector of gradients of loss function.
It is defined as a notation compressed with  $\mathbf{g}_{1:t}=\sum_{s=1}^t \mathbf{g}_{s}$. Recall that $g_{s, i}$ is the $i^{th}$ batch coordinate of the gradient $\mathbf{g}_{s} = \nabla \ell_{s}(\mathbf{w}_{s})$. Then, Learning rate $\eta_{t}$ equation proposed in ~\cite{McMahan:2013:ACP:2487575.2488200} as \begin{equation} \eta_{t, i} = \frac{\alpha}{\beta + \sqrt{{\sum_{s=1}^t}g_{s, i}^2}},  \end{equation}
where $\alpha$ is set as twice the maximum allowed magnitude for $\mathbf{w}_{i}$ to give the best possible regret bound. $\beta$ = 1 is usually good enough. This simply ensures that early learning rates are not too high~\cite{McMahan:2013:ACP:2487575.2488200}.

The first term $\mathbf{g}_{1:t}\cdot \mathbf{w}$ expresses the FTL(Follow the leader) and takes a leader that has the smallest loss until $t$ step to obtain an approximation of the loss. The second term $\frac{1}{2}\sum_{s=1}^t \sigma_{s}||\mathbf{w}-\mathbf{w}_{s}||^2_2$ is a proximal part that limits the new weight to no large fluctuations in the weights previously used. Increases the stability of the algorithm by adding strong convexity. The third term $\lambda_1 ||\mathbf{w}||_1$ is the L1 regularization part. 

If there is no distortion or bias, the more data there is in machine learning, the more accurate and robust the prediction model is.  However, the media market continues to feature new types of content, and new actors and new program material are frequently registered. These factors continue to give us the volatility of the keyword data. It is difficult to understand all of these volatility as the model learns. We effectively overcome this problem by using the FTRL-proximal algorithm. The experimental results for this are shown in section 4.

 \subsection{Classification Definition of contents to be published}

  Since we divide into two categories of dataset and then classify the contents as popular contents, we need to establish the interface definitions between these two models so that we can present the final popular contents classification probability. First, we predict popular probability $P_{c_{n}\in A}$ as a Boosting model for the characteristic $A$ dataset that accounts for more than 70\% of $c_{n}(=c_{new})$. Next, calculate the $P_{c_{n}\in B}$ through the deep neural network in the remaining characteristic $B$ dataset. Indeed, popular contents are concentrated in an characteristic $A$ dataset, and the ratio of the data is much higher than that of characteristic $B$ dataset. In addition, the performance of our model, which will be described later, exceeds 90\%, so even if we predict  $P_{c_{n}\in B}$ after predicting  $P_{c_{n}\in A}$, we will almost converge to the popular contents boundary that we already defined. Of course, we can sort the view count values of two datasets by regression at the same time to perform storage allocation. However, predicting the unpublished contents through regression is very limiting in terms of performance issues because the lack of historical data makes it impossible to grasp the time series. Finally, we use Eq.1 to calculate the final predicted probability of content $c_{n, i}$($i$ is current content index) at time $t$ using the information(contents $c_{old}$ already released) available until $t-r$.  As described above, since the whether or not to use of text data depends on the characteristics of the dataset, we set the period $r$ differently considering the volatility of the text data.

\begin{equation*} 
\widehat{P}_{c_{n, i}}(t-r,t) = \begin{cases}
\widehat{P}_{{c}_{n, i}}(t-r_{{}_{A}},t), & \mbox{if}i \in A set \\
\widehat{P}_{{c}_{n, i}}(t-r_{{}_{B}},t), & \mbox{if}i \in B set  \end{cases} 
\end{equation*}

\section{Experiments}
\subsection{Dataset}
The dataset in the experiment is taken from one of the top Korean online streaming video service companies. This company produces more than 50 million data logs every month.  We collect the dataset including the metadata and historical view-count of the contents from the service. Content that has existed before and whose rating can be tracked is not subject to our analysis, and new incoming content is only our analysis target at T present time.  The  features provided by the service company in advance are composed of 10 variables except sequel episode count and Related view.  The dataset does not contain a feature based on the view-count record as log data and is the contents metadata before the contents is released. Although we use the dataset containing view-count log to classify whether the content is hot or cold when we train model, we do not have any view-count log when predicting the new release content.

We will also add view-count for related episode content to variables. This variable is the only static data we have, and later experiments have shown that it is the most important variable. This process of extracting as much of the historical data as possible from the time series perspective of the data is critical for the view count prediction experiment. However, content that is first published in platform or movie contents is limited to extracting variables for these related works. First, we divide the data into two clusters based on the availability of the historical data. A cluster that has access to historical data uses the XGBoosting technique, and a cluster that does not use historical data uses the neural net with embedding that also uses the unstructured data, such as text data to overcome it. Finally, models will categorize hot and cold contents.

\subsection{Experimental setting}

\subsubsection{Experiment environment:} 
In this study, we use Python-based machine learning framework to achieve the results. we use Scikit-Learn\footnote{https://scikit-learn.org/}, fast.ai\footnote{https://docs.fast.ai/}, and Keras~\cite{chollet2018keras} which open source neural network library. The machine we used has 64 Gigabyte memory and GTX 1080 with 8 Gigabyte GPU memory.

\subsubsection{Parameters Setting:} Keras is used to build our neural net model. The batch size is set to 64 and the epoch is set 50. We train the model with  the hard-sigmoid activation function. Vector size for categorical data embeddings is set to 30.

\subsection{Prediction Results}
Our experimental results are evaluated by comparing the predictions of our model and the actual classification of the content.  Our current data does not have data from previous years prior to January, so there are not enough contents to qualify new contents in January and February.  Therefore, data from March to December 2017 are used in the experiment. We evaluate the performance with precision, recall, and f1-score and compare it with the techniques previously utilized to predict the classification of contents popularity.  The parameters of the model are empirically found by optimizing the results of the validation set. 

Table \ref{tab:b} shows the results of the comparison with the Random Forest, XGBoosting, SVR~\cite{suykens1999least}, and neural net models used in similar studies. The performance of our model is higher than that of other existing models to all performance metrics in 15,376 cumulative new publish contents for nine months.

\begin{table}[!h]
\centering
\caption{Comparison of Performance for BOW-SVR, RF, XGBOOST and Our Model}
\label{tab:b}
\setlength{\tabcolsep}{11pt}
\renewcommand{\arraystretch}{1.2}
\begin{tabular}{|c|c|c|c|}
\hline
Numbers of Programs & \multicolumn{3}{c|}{15,376}                      \\ \hline
Method              & Precision      & Recall         & F1-Score       \\ \hline
RF                  & 0.951          & 0.737          & 0.830          \\
XGB                 & 0.942          & 0.743          & 0.842          \\
SVR                 & 0.892          & 0.529          & 0.738          \\
MLP                 & 0.889          & 0.772          & 0.826          \\ \hline
Our Model           & \textbf{0.952} & \textbf{0.852} & \textbf{0.895} \\ \hline
\end{tabular}
\end{table}

Table \ref{tab:c} shows the performance comparison between the categorical variables using embedding and one-hot encoding. Precision, Recall and f1-score all have high performance in models with categorical data embedding technique. We can see that the highest performance improvement achieved by categorical embedding is Neural Network.

\begin{table}[!h]
\centering
\caption{Performance comparison by model with or without Categorical Embedding.}
\label{tab:c}
\setlength{\tabcolsep}{12pt}
\renewcommand{\arraystretch}{1.3}
\begin{tabular}{|c|c|c|}
\hline
Method    & F1-Score       & F1-Score(using categorical embedding) \\ \hline
RF        & 0.822          & 0.830              \\
XGB       & 0.821          & 0.842              \\
SVR       & 0.682        & 0.738              \\
MLP       & 0.774          & 0.826              \\ \hline
Our Model & \textbf{0.851} & \textbf{0.895}     \\ \hline
\end{tabular}
\end{table}

\begin{figure}
\centering
\begin{tikzpicture}
\begin{axis}[
    xlabel={Epoch},
    ylabel={Loss},
    xmin=0, xmax=50,
    ymin=0.1, ymax=0.6,
    xtick={0,10,20,30,40,50},
    ytick={0,0.2,0.4,0.6},
    legend pos=north east,
    ymajorgrids=true,
    grid style=dashed,
]
 
\addplot[
    color=blue
    ]
    coordinates {
    (0,1.4584)(1,0.6714)(5,0.3639)(10,0.3003)(15,0.2699)(20,0.2559)(25,0.2615)(30,0.2616)(35,0.2500)(40,0.2347)(45,0.2322)(50, 0.2273)
    };
    
\addplot[
    color=red
    ]
    coordinates {
    (0,1.5)(1,1.4391)(5,0.4648)(10,0.3132)(15,0.3006)(20,0.2807)(25,0.2669)(30,0.2707)(35,0.2513)(40,0.2464)(45,0.2482)(50, 0.2559)
    };
    
\addplot[
    color=green
    ]
    coordinates {
    (0,1.5242)(1,0.6302)(5,0.4528)(10,0.3519)(15,0.3053)(20,0.2910)(25,0.2900)(30,0.2971)(35,0.2759)(40,0.2935)(45,0.2897)(50, 0.2978)
    };
    
\addplot[
    color=yellow
    ]
    coordinates {
    (0,1.5584)(1,0.7714)(5,0.3939)(10,0.3233)(15,0.3018)(20,0.3185)(25,0.3018)(30,0.3125)(35,0.3098)(40,0.2982)(45,0.2957)(50, 0.2876)
    };
    
\addlegendentry{Ftrl-Proximal}
\addlegendentry{Adam}
\addlegendentry{RMSprop}
\addlegendentry{FOBOS}

\end{axis}
\end{tikzpicture}
\caption{Comparison of Training Loss for various optimizer method.}
\label{fig:c}
\end{figure}
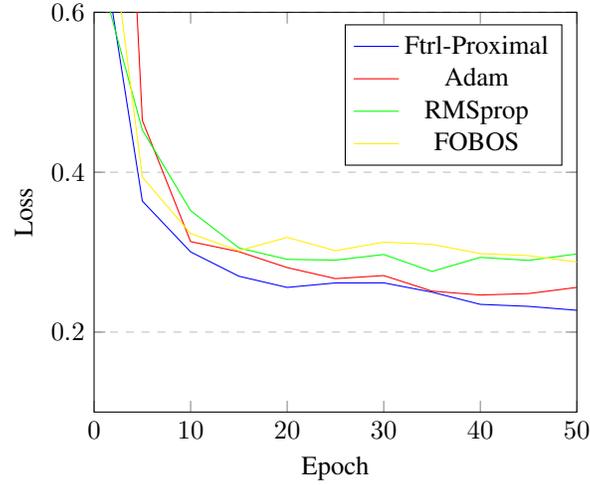

Fig~\ref{fig:c} shows the convergence results on media streaming dataset. The results show that Ftrl-proximal Optimizer outperforms Other optimization method such as RMSprop, Adam ~\cite{kingma2014adam}, FOBOS ~\cite{duchi2009efficient} by around 5.6\% on average. These three algorithms are almost indistinguishable in the their trade off curves on the our dataset, but on the FTRL-Proximal learned faster and showed better performance.

\begin{figure}[!h]
\centering
\begin{tikzpicture}
\begin{axis}[
    xlabel={Observation period},
    ylabel={F1-Score},
    xmin=0, xmax=40,
    ymin=0.4, ymax=1.0,
    xtick={0,10,20,30,40},
    ytick={0.4, 0.5, 0.6,0.7,0.8,0.9,1.0},
    legend pos=south east,
    ymajorgrids=true,
    grid style=dashed,
]
 
\addplot[
    color=blue
    ]
    coordinates {
    (0,0)(1,0.6852)(5,0.8163)(10,0.8654)(15,0.8795)(20,0.8879)(25,0.8926)(30,0.8996)(35,0.8906)(40,0.8874)(45,0.8759)(50, 0.8579)
    };
    
\addplot[
    color=red
    ]
    coordinates {
    (0,0)(1,0.4852)(5,0.6268)(10,0.6878)(15,0.7092)(20,0.7598)(25,0.7497)(30,0.7648)(35,0.7518)(40,0.7395)(45,0.7468)(50, 0.7490)
    };
    
\addlegendentry{Type {A}}
\addlegendentry{Type{B}}

\end{axis}
\end{tikzpicture}
\caption{Comparison of F1-Score as a function of Observation Period for various Model.}
\label{fig:d}
\end{figure}
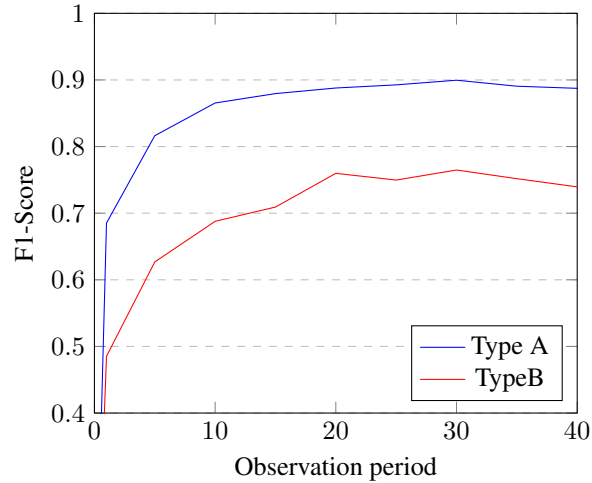

Fig~\ref{fig:d} shows the f1-score values produced with different contents observation period r to be used as training data by all 4 models. As we have already mentioned in section 3 above, we designated different observation periods depending on the characteristics of the two data. In the case of type A, performance will not rise as long as the observation period exceeds 10 days. Likewise, type B no longer rises after 20 days. We have defined 10 and 20 for $r_a$ and $r_b$, respectively, as defined in Eq.4 above, and we were able to achieve the best performance and efficient computation time in the experiment.

\section{Conclusion \& Future work}
In this paper, we suggest a hybrid model to the popularity prediction of newly video contents. We divide two types of contents, select important feature for sequal contents, and apply two models, which are XGBoost and deep neural net, to two data types. To complement the shortage of features to standalone contents, the untrusted data such as text are utilized with structured data, and embedding techniques are adopted to train both data together. To consider characteristics of the view-count Volatility of streaming video contents, we apply the FTRL-Proximal optimizer to the gradient-based learning process of the proposed models. Finally, we experiment the hybrid model to the real dataset using one of the top streaming companies in Korea, and achieving overall better performance compared to the existing standalone models.

However, there are some challenges that we have not yet tried to improve upon in the performance of our model. Before the program was added to the online streaming service, there are important external data that are not considered in our model, such as the number of viewers and audience for each program and movie, movie rating, reputation of director and actor, and emotional analysis data of the public. A text-based analysis method based on the viewer's reputation for content can be used to improve the accuracy of the model in future work. Although this research is a significant contribution to the early stages of research by only focusing on the popularity prediction of the new contents not published, there are numerous challenges that must be addressed to improve the prediction performance. Before the program was added to the online streaming service, there were important external data that were not considered in the proposed model, such as the number of viewers and audience for movies, movie rating, the reputation of director and actor, and emotional analysis data of the public. A text-based analysis method based on the viewer's reputation for content can be used to improve the accuracy of the model in future work. First of all, a well-organized benchmarking dataset of streaming video contents is needed. Because of privacy issue, it's very hard to open the company's dataset to the public, however, s discussion in the academic community with a commercial organization is needed.

\bibliographystyle{unsrt}

\begin{thebibliography}{10}

\bibitem{7047922}
L.~Chen, Y.~Zhou, and D.~M. Chiu.
\newblock Smart streaming for online video services.
\newblock {\em IEEE Transactions on Multimedia}, 17(4):485--497, April 2015.

\bibitem{McMahan:2013:ACP:2487575.2488200}
H.~Brendan McMahan, Gary Holt, D.~Sculley, Michael Young, Dietmar Ebner, Julian
  Grady, Lan Nie, Todd Phillips, Eugene Davydov, Daniel Golovin, Sharat
  Chikkerur, Dan Liu, Martin Wattenberg, Arnar~Mar Hrafnkelsson, Tom Boulos,
  and Jeremy Kubica.
\newblock Ad click prediction: A view from the trenches.
\newblock In {\em Proceedings of the 19th ACM SIGKDD International Conference
  on Knowledge Discovery and Data Mining}, KDD '13, pages 1222--1230, New York,
  NY, USA, 2013. ACM.

\bibitem{chen2016xgboost}
Tianqi Chen and Carlos Guestrin.
\newblock Xgboost: A scalable tree boosting system.
\newblock In {\em Proceedings of the 22nd acm sigkdd international conference
  on knowledge discovery and data mining}, pages 785--794. ACM, 2016.

\bibitem{DBLP:journals/corr/ChenG16}
Tianqi Chen and Carlos Guestrin.
\newblock Xgboost: {A} scalable tree boosting system.
\newblock {\em CoRR}, abs/1603.02754, 2016.

\bibitem{MEDIA}
Jingsong~Cui Scott~Sereday.
\newblock Using machine learning to predict future tv ratings.
\newblock {\em nielsen journal}, 1(3), February 2017.

\bibitem{8086153}
C.~Zhu, G.~Cheng, and K.~Wang.
\newblock Big data analytics for program popularity prediction in broadcast tv
  industries.
\newblock {\em IEEE Access}, 5:24593--24601, 2017.

\bibitem{kmeans}
B.RAVEENA D.~ANAND, A.V.SATYAVANI and M.POOJITHA.
\newblock Analysis and prediction of television show popularity rating using
  incremental k-means algorithm.
\newblock {\em International Journal of Mechanical Engineering \& Technology
  (IJMET)}, 9(1):482--489, January 2018.

\bibitem{7545038}
Y.~Fukushima, T.~Yamasaki, and K.~Aizawa.
\newblock Audience ratings prediction of tv dramas based on the cast and their
  popularity.
\newblock In {\em 2016 IEEE Second International Conference on Multimedia Big
  Data (BigMM)}, pages 279--286, April 2016.

\bibitem{10.1371/journal.pone.0071226}
Márton Mestyán, Taha Yasseri, and János Kertész.
\newblock Early prediction of movie box office success based on wikipedia
  activity big data.
\newblock {\em PLOS ONE}, 8(8):1--8, 08 2013.

\bibitem{DBLP:journals/corr/abs-1301-3781}
Tomas Mikolov, Kai Chen, Greg Corrado, and Jeffrey Dean.
\newblock Efficient estimation of word representations in vector space.
\newblock {\em CoRR}, abs/1301.3781, 2013.

\bibitem{Bartlett:2007:AOG:2981562.2981571}
Peter~L. Bartlett, Elad Hazan, and Alexander Rakhlin.
\newblock Adaptive online gradient descent.
\newblock In {\em Proceedings of the 20th International Conference on Neural
  Information Processing Systems}, NIPS'07, pages 65--72, USA, 2007. Curran
  Associates Inc.

\bibitem{chollet2018keras}
Fran{\c{c}}ois Chollet et~al.
\newblock Keras: The python deep learning library.
\newblock {\em Astrophysics Source Code Library}, 2018.

\bibitem{suykens1999least}
Johan~AK Suykens and Joos Vandewalle.
\newblock Least squares support vector machine classifiers.
\newblock {\em Neural processing letters}, 9(3):293--300, 1999.

\bibitem{kingma2014adam}
Diederik~P Kingma and Jimmy Ba.
\newblock Adam: A method for stochastic optimization.
\newblock {\em arXiv preprint arXiv:1412.6980}, 2014.

\bibitem{duchi2009efficient}
John Duchi and Yoram Singer.
\newblock Efficient online and batch learning using forward backward splitting.
\newblock {\em Journal of Machine Learning Research}, 10(Dec):2899--2934, 2009.

\end{thebibliography}

\end{document}